\documentclass{article}

\usepackage[preprint]{neurips_2025}

\usepackage[utf8]{inputenc}
\usepackage[T1]{fontenc}
\usepackage{hyperref}
\usepackage{url}
\usepackage{booktabs}
\usepackage{amsfonts}
\usepackage{nicefrac}
\usepackage{microtype}
\usepackage{graphicx}
\usepackage{amsmath,amssymb}
\usepackage{math_commands}
\usepackage{xcolor}
\usepackage{algorithm}
\usepackage{algorithmic}
\usepackage{multirow}
\usepackage{placeins}
\usepackage{float}

\title{NOBLE: Accelerating Transformers with\\Nonlinear Low-Rank Branches}

\author{%
  Ethan Smith \\
  Canva Research \\
  \texttt{ethansmith@canva.com}
}

\begin{document}
\raggedbottom

\maketitle
\vspace{-1.5em}

\begin{abstract}
We introduce \textbf{NOBLE} (\textbf{N}onlinear l\textbf{O}w-rank \textbf{B}ranch for \textbf{L}inear \textbf{E}nhancement), an architectural augmentation that adds nonlinear low-rank branches to transformer linear layers. Unlike LoRA and other parameter-efficient fine-tuning (PEFT) methods, NOBLE is designed for pretraining from scratch. The branch is a permanent part of the architecture as opposed to an adapter for finetuning on top of frozen weights. The branch computes $\sigma(xW_\text{down})W_\text{up}$ where $\sigma$ is a learnable nonlinearity. We evaluate several activation functions and find that \textbf{CosNet}, a two-layer cosine nonlinearity with learnable frequency and phase with a linear projection in between them in the bottleneck space, performs best. NOBLE achieves substantial improvements with minimal overhead: \textbf{up to 1.47$\times$ step speedup} to reach baseline eval loss (up to 32\% fewer training steps), with as low as 4\% additional parameters and 7\% step time overhead, resulting in \textbf{up to 1.22$\times$ net wallclock speedup}. Experiments on LLMs (250M and 1.5B parameters), BERT, autoregressive image token modeling, and ViT consistently show improved training efficiency. We identify one caveat: Mixup/CutMix augmentation interferes with NOBLE's benefits in Imagenet classification along with other stochastic augmentations, but when disabled, ViT also improves. This discrepancy is possibly explained by regularization techniques that encourage smoother fits to the target function while NOBLE may specialize more in sharper aspects of the target function.
\end{abstract}

\vspace{-2.0em}
\begin{figure}[H]
\centering
\includegraphics[width=0.87\textwidth]{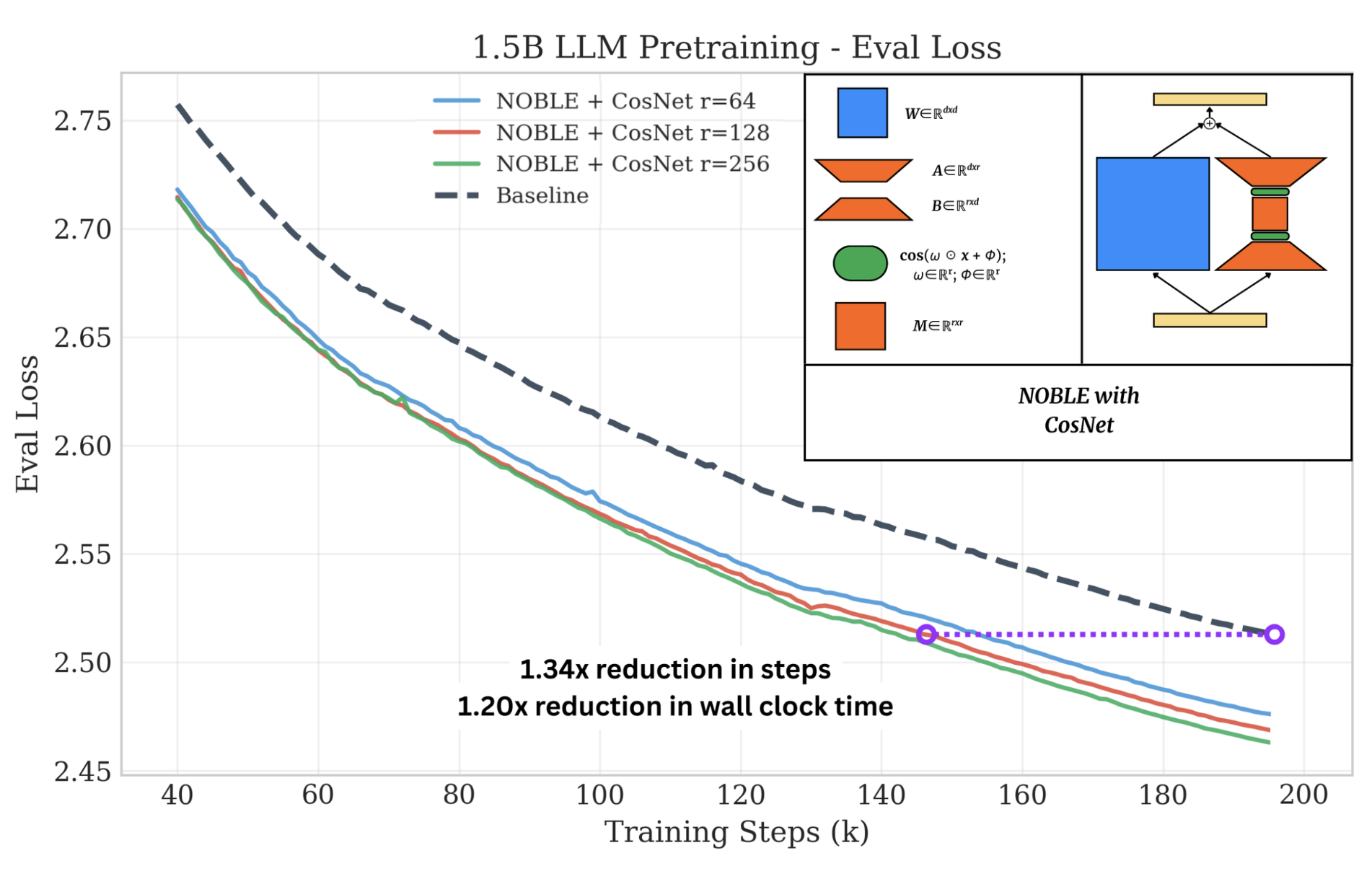}
\caption{\textbf{Eval loss curves for 1.5B LLM pretraining on OpenWebText.} NOBLE (solid) reaches baseline's best loss (2.51) in 143--154k steps vs 196k (1.34--1.37$\times$ faster). First 20\% of training truncated.}
\label{fig:hero}
\end{figure}

\section{Introduction}

Transformers have become the dominant architecture for both natural language processing~\citep{vaswani2017attention,brown2020language} and computer vision~\citep{dosovitskiy2020image}. Despite their success, the linear projections that make up the bulk of transformer parameters, those in attention layers and feedforward networks, are fundamentally limited to computing affine transformations within each layer. While nonlinearities in feedforward blocks provide expressivity, the attention mechanism's query, key, and value projections remain purely linear.

Low-Rank Adaptation (LoRA)~\citep{hu2021lora} demonstrated that fine-tuning large models can be achieved by learning low-rank additive updates to frozen weights. However, naively applying linear LoRA to pretraining offers limited benefit: the low-rank matrices $BA$ can be fused into the main weight matrix $W + BA$, collapsing to a standard linear layer with slightly different initialization. Without a nonlinearity, the ``bypass'' may not function as a separate computational branch but instead closer to a delta to the main weights. This raises a natural question: \emph{can we design low-rank branches with true architectural benefits during pretraining, not just parameter-efficient fine-tuning?}

We introduce \textbf{NOBLE} (\textbf{N}onlinear l\textbf{O}w-rank \textbf{B}ranch for \textbf{L}inear \textbf{E}nhancement), a simple modification that augments linear layers with a nonlinear low-rank branch. \textbf{Crucially, unlike LoRA and the broader family of Parameter-Efficient Fine-Tuning (PEFT) methods, NOBLE is not a fine-tuning technique. It is a fundamental architectural augmentation designed for pretraining~from~scratch.} The low-rank branch becomes a permanent part of the model architecture, trained jointly with all other parameters from initialization.

The key insight is that a small bottleneck with a well-chosen nonlinearity can capture function variations that complement the main linear pathway. We evaluate several activation functions and find that cosine-based activations perform best, with our recommended variant \textbf{CosNet}, two cosine nonlinearities with learnable frequency and phase along with a linear projection in between them in the bottleneck space, achieving the strongest results. Our approach differs from standard LoRA in three important ways:
\begin{enumerate}
    \item \textbf{Architectural augmentation, not PEFT}: The branch is a permanent part of the architecture trained from scratch, not an adapter added to frozen weights for fine-tuning~\citep{houlsby2019parameter}.
    \item \textbf{Nonlinear activation}: We evaluate several activations and recommend CosNet: two cosine nonlinearities with learnable frequency and phase, connected by a small mixing matrix.
    \item \textbf{Learning rate scaling}: We apply elevated learning rates to $W_\text{up}$ and $M$ based on the ratio $(\text{dim}/r)^\gamma$, roughly following insights from $\mu$P~\citep{yang2022tensor}.
\end{enumerate}

\textbf{Main result: up to 1.47$\times$ training speedup on language models.} Our experiments demonstrate that NOBLE dramatically improves training efficiency for language models:
\begin{itemize}
    \item \textbf{21--32\% fewer steps}: Models reach baseline eval loss in up to 1.47$\times$ fewer training steps (e.g., 169k vs 249k steps for rank 256).
    \item \textbf{Modest overhead}: 4--24\% additional parameters and 7--21\% longer step times.
    \item \textbf{Net wallclock savings}: Despite the per-step overhead, the reduced step count yields 1.17--1.22$\times$ wallclock speedup.
    \item \textbf{Better final loss}: At convergence, NOBLE achieves 0.02--0.07 lower eval loss than baseline in the experiments we conducted.
\end{itemize}

We validate these results across LLM autoregressive training at two scales (250M and 1.5B parameters), 110M BERT-style masked language modeling, ViT-Small ImageNet classification, and autoregressive image token modeling on ImageNet, with ablations on rank (64, 128, 256).

\textbf{Contributions}
\begin{itemize}
    \item We propose NOBLE, a family of nonlinear low-rank branches for linear layer enhancement, achieving \textbf{up to 1.47$\times$ step speedup} and \textbf{1.17--1.22$\times$ wallclock speedup} on autoregressive pretraining with minimal overhead.
    \item We evaluate several activation functions and identify \textbf{CosNet}, a sandwich of a learnable cosine activation, a linear projection, and then another learnable cosine activation, as the best-performing variant.
    \item We introduce key design choices: near-zero initialization and scaled learning rates for low-rank components.
    \item We provide extensive experiments on LLMs (two scales), BERT, ViT, and autoregressive image token modeling, demonstrating broad effectiveness across tasks, with a note that aggressive augmentation (Mixup/CutMix) can interfere with NOBLE's benefits.
\end{itemize}

\section{Related Work}

\paragraph{Low-Rank Adaptation and PEFT.} LoRA~\citep{hu2021lora} introduced low-rank updates for parameter-efficient fine-tuning (PEFT), spawning a large family of methods. LoRA+~\citep{hayou2024loraplus} showed that using different learning rates for the down and up projection matrices improves fine-tuning efficiency, we adopt a similar asymmetric learning rate strategy for NOBLE. DoRA~\citep{liu2024dora} decomposes weights into magnitude and direction components. AdaLoRA~\citep{zhang2023adalora} learns adaptive ranks, and QLoRA~\citep{dettmers2023qlora} enables fine-tuning with quantized weights. MoSLoRA~\citep{wu2024moslora} introduces mixture-of-subspaces to improve LoRA's expressiveness. These methods share a common paradigm: freeze pretrained weights and learn low-rank \emph{corrections} for downstream tasks. NOBLE takes a fundamentally different approach, rather than a fine-tuning adapter, we propose a \emph{permanent architectural augmentation} trained from scratch during pretraining.

\begin{figure}[H]
\centering
\includegraphics[width=\textwidth]{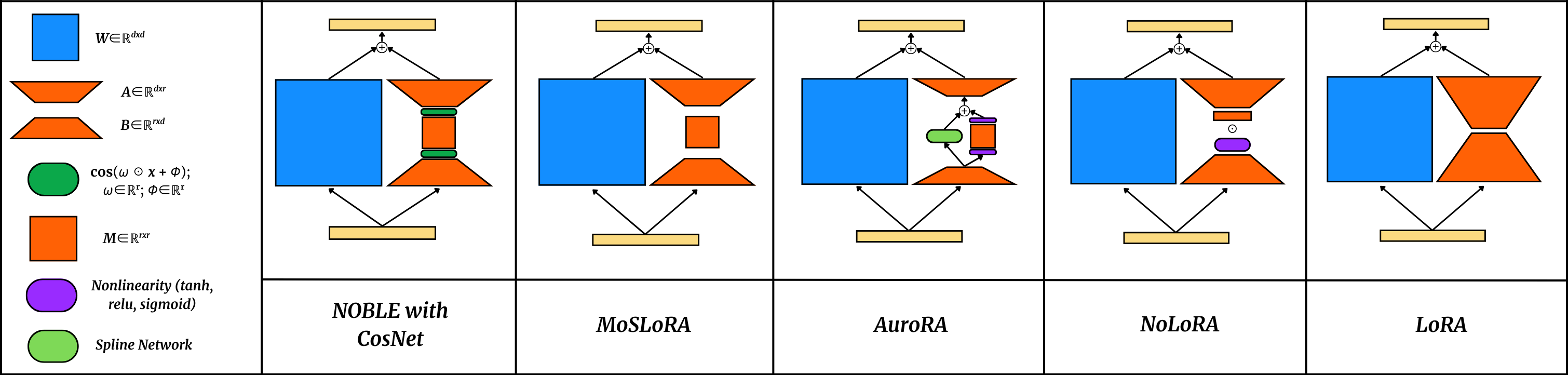}
\caption{\textbf{Comparison of low-rank adaptation architectures.} (a) NOBLE with CosNet: a sandwich of cosine activations with learnable frequency/phase, connected by a mixing projection, before projecting back to full dimensions. Key differences: trained from scratch as part of the architecture (not fine-tuning), cosine activation is symmetric and non-saturating. (b) MosLoRA adds a linear projection between low-rank matrices $BMA$, similar structure but without nonlinear activations. (c--d) Recent nonlinear variants (AuroRA, NoLoRA) introduce activations in the LoRA bottleneck for fine-tuning. (e) Standard LoRA adds a linear low-rank bypass $BA$.}
\label{fig:arch_comparison}
\end{figure}

\paragraph{Nonlinear Extensions of LoRA.} Several concurrent works have explored adding nonlinearity to LoRA for fine-tuning (see Figure~\ref{fig:arch_comparison} for architectural comparison). NEAT~\citep{zhong2024neat} learns a nonlinear transformation of pretrained weights to approximate cumulative updates. AuroRA~\citep{dong2024aurora} incorporates an Adaptive Nonlinear Layer between linear projectors to capture fixed and learnable nonlinearities. These methods target the PEFT setting (fine-tuning frozen models), whereas NOBLE is designed for pretraining from scratch (although given the similarity in architecture to related works, it may also be used for fine-tuning as well, or possibly continued pretraining for further reduction in~loss). Additionally, our work identifies that cosine activations are particularly effective in low-rank bottlenecks due to their symmetry and non-saturating properties.

\paragraph{Nonlinear Projections in Attention.} Several works have explored adding nonlinearity to attention mechanisms~\citep{choromanski2020rethinking,katharopoulos2020transformers}. \citet{so2021primer} searched for architectural improvements including activation functions. Our approach is orthogonal, adding nonlinearity via bypass branches rather than modifying the main computation.

\paragraph{Mixture of Experts.} MoE layers~\citep{shazeer2017outrageously,fedus2022switch} add capacity through conditional computation. NOBLE provides a simpler, dense alternative that adds capacity through low-rank nonlinear branches without routing overhead.

\paragraph{Periodic Activations.} Sinusoidal representations have proven effective in neural radiance fields~\citep{mildenhall2020nerf} and implicit neural representations~\citep{sitzmann2020implicit}. We adopt learnable cosine activations in our bypass.

\section{Method}

\subsection{NOBLE Architecture}

Given a standard linear layer $f(x) = xW + b$ with $W \in \R^{d_\text{in} \times d_\text{out}}$, NOBLE augments it with a low-rank nonlinear branch:
\begin{equation}
    f_\text{NOBLE}(x) = xW + b + \sigma(xW_\text{down})W_\text{up}
\end{equation}
where $W_\text{down} \in \R^{d_\text{in} \times r}$, $W_\text{up} \in \R^{r \times d_\text{out}}$, $r \ll \min(d_\text{in}, d_\text{out})$ is the bottleneck rank, and $\sigma$ is a nonlinear activation function. We evaluated several choices for $\sigma$ and find that cosine-based activations perform best (see Section~\ref{sec:experiments}).

\subsection{CosNet: Recommended Nonlinearity}

Among several activation functions we evaluate (see Section~\ref{sec:experiments}), we recommend \textbf{CosNet}, a two-layer cosine nonlinearity operating in the bottleneck space:
\begin{equation}
    \sigma_\text{cos}(h) = \cos(\omega_2 \odot (M \cdot \cos(\omega_1 \odot h + \phi_1)) + \phi_2)
\end{equation}
where $h \in \R^r$ is the bottleneck representation, $M \in \R^{r \times r}$ is a learned mixing matrix, and each dimension $i$ has learnable frequency $\omega_i$ and phase $\phi_i$ parameters.

The cosine activation $\cos(\omega \odot x + \phi)$ has several properties we theorize may be desirable:
\begin{itemize}
    \item \textbf{Bounded output}: Unlike ReLU or GELU, cosine outputs are bounded in $[-1, 1]$, providing natural regularization.
    \item \textbf{Smooth and periodic}: Enables modeling of periodic patterns and smooth interpolations.
    \item \textbf{Learnable frequency}: The frequency parameters $\omega$ control the sensitivity to input variations, adapting to the data distribution.
    \item \textbf{Learnable phase}: The phase parameters $\phi$ allow shifting the nonlinearity's operating point.
\end{itemize}

\subsection{Key Design Choices}

\paragraph{Near-Zero Initialization of $W_\text{up}$.} We initialize $W_\text{up}$ with small standard deviation $\alpha / \sqrt{r}$ with $\alpha = 0.01$, ensuring the branch contributes negligibly at initialization. This allows the main linear layer to dominate early training while the branch gradually learns complementary features.

\paragraph{Reduced Main Weight Initialization.} We initialize the main linear layer $W$ with standard deviation $0.5 / \sqrt{d_\text{in}}$, half the typical Kaiming scale. This slightly reduces the main pathway's initial magnitude, leaving more room for the branch to contribute as training progresses.

\paragraph{Learning Rate Scaling.} The low-rank structure necessitates careful learning rate treatment. Roughly following $\mu$P~\citep{yang2022tensor}, we apply elevated learning rates to $W_\text{up}$ and the CosNet mixing matrix $M$, while $W_\text{down}$ uses the base learning rate. Specifically:
\begin{align}
    \text{lr}_{W_\text{up}} &= \text{lr}_\text{base} \cdot \left(\frac{\min(d_\text{in}, d_\text{out})}{r}\right)^{2\gamma} \\
    \text{lr}_{M} &= \text{lr}_\text{base} \cdot \left(\frac{\min(d_\text{in}, d_\text{out})}{r}\right)^{\gamma_M}
\end{align}
where $\gamma = 0.3$ is the base learning rate multiplier power (so $W_\text{up}$ uses exponent $2\gamma = 0.6$), and $\gamma_M = 0.45$ for the mixing matrix.

\paragraph{CosNet Initialization.} We initialize the learnable frequency multipliers by sampling uniformly from $[\omega_\text{min}, \omega_\text{max}]$ (default: $[0.8, 1.2]$). Phases are initialized from $\mathcal{N}(0, \sigma_\phi^2)$ with $\sigma_\phi = 0.1$. The mixing matrix $M$ uses Xavier initialization.

\subsection{Computational Overhead}

For typical settings ($d_\text{in} = d_\text{out} = d$, $r = 64$--$128$), NOBLE adds:
\begin{itemize}
    \item \textbf{Parameters}: 4--12\% additional (low-rank matrices + CosNet parameters)
    \item \textbf{Step time}: 7--12\% longer (measured on H100 GPUs using with bf16 mixed precision, torch.compile using reduce-overhead mode and fullgraph=True)
\end{itemize}
These percentages are \emph{relative to model size}: the same rank on a larger model yields proportionally smaller overhead. For example, rank 64 adds 5.7\% parameters to our 250M base model but only 4.0\% to our 1.5B large model (see Table~\ref{tab:lm_results}). Critically, despite this per-step overhead, the 21--32\% reduction in training steps yields \textbf{net wallclock speedups of 1.17--1.22$\times$}.

\section{Experiments}
\label{sec:experiments}

We evaluate NOBLE across five experiments spanning language and vision tasks. Our central finding is that NOBLE provides consistent and significant benefits for autoregressive token prediction (LLM, image tokens) and masked language modeling (BERT), but does not improve ViT image classification when certain augmentations are enabled in training.

\subsection{Language Model Experiments}

\subsubsection{Autoregressive LLM Pretraining}

\textbf{Setup.} We train LLM transformers on OpenWebText~\citep{gokaslan2019openwebtext}. The architecture uses GeGLU feedforward blocks~\citep{shazeer2020glu} with hidden dimension $\frac{8}{3}d$ (aligned to 64), RoPE positional encodings~\citep{su2024roformer}, and RMSNorm~\citep{zhang2019root}. We test two configurations:
\begin{itemize}
    \item \textbf{Base}: depth 12, width 1024, 8 attention heads, batch 64, lr $3 \times 10^{-4}$, 250k steps
    \item \textbf{Large}: depth 24, width 2048, 16 attention heads, batch 128, lr $2.5 \times 10^{-4}$, 200k steps
\end{itemize}
Both use weight decay 0.01, $\beta_1=0.9$, $\beta_2=0.98$, sequence length 1024.

NOBLE is applied to all linear projections (Q, K, V, output projection, and both FFN layers, we leave ablations on selecting which layers to apply NOBLE to for future work). We ablate across ranks 64, 128, and 256.

\begin{figure}[t]
\centering
\begin{minipage}{0.48\textwidth}
    \centering
    \includegraphics[width=\textwidth]{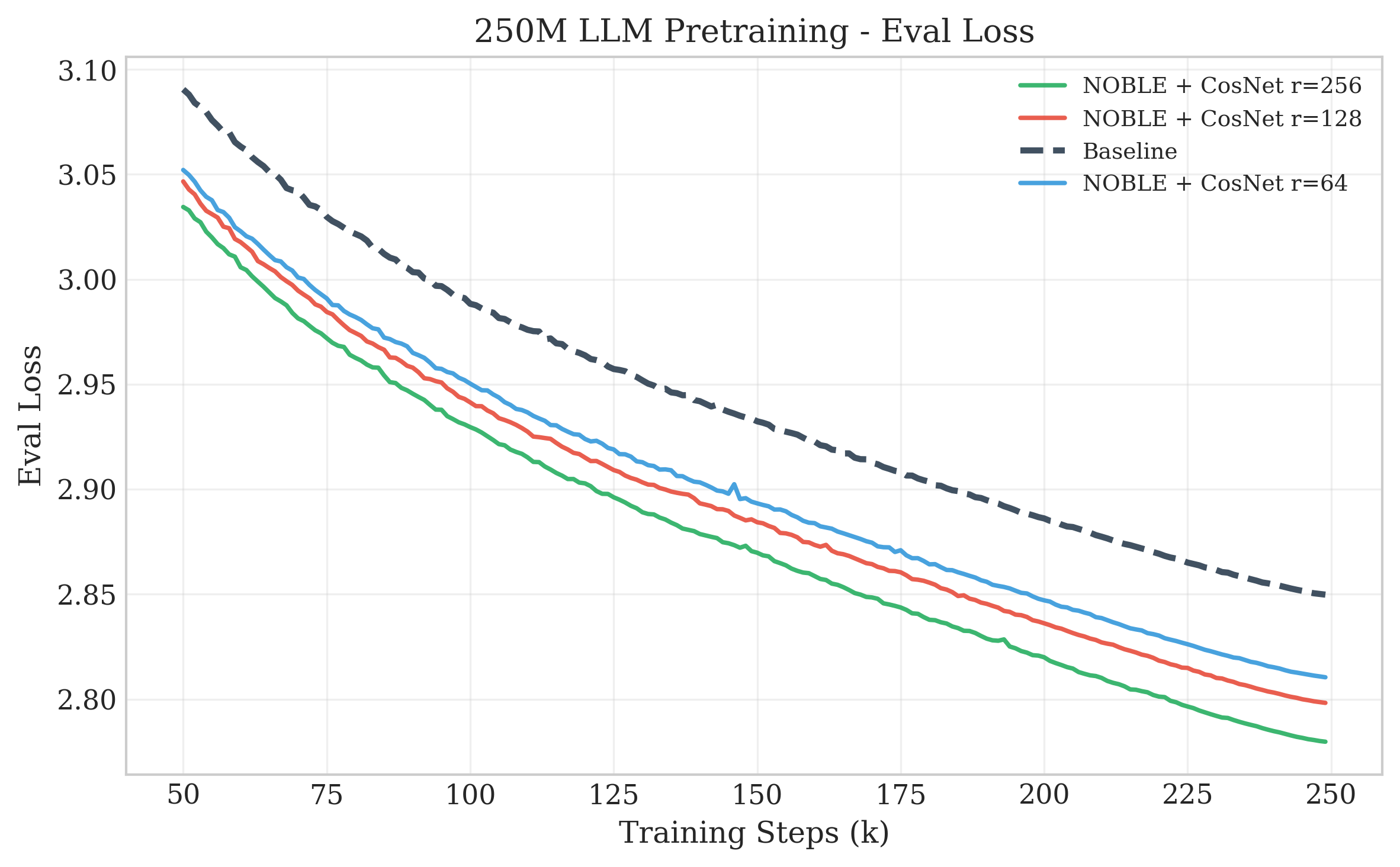}
\end{minipage}
\hfill
\begin{minipage}{0.48\textwidth}
    \centering
    \includegraphics[width=\textwidth]{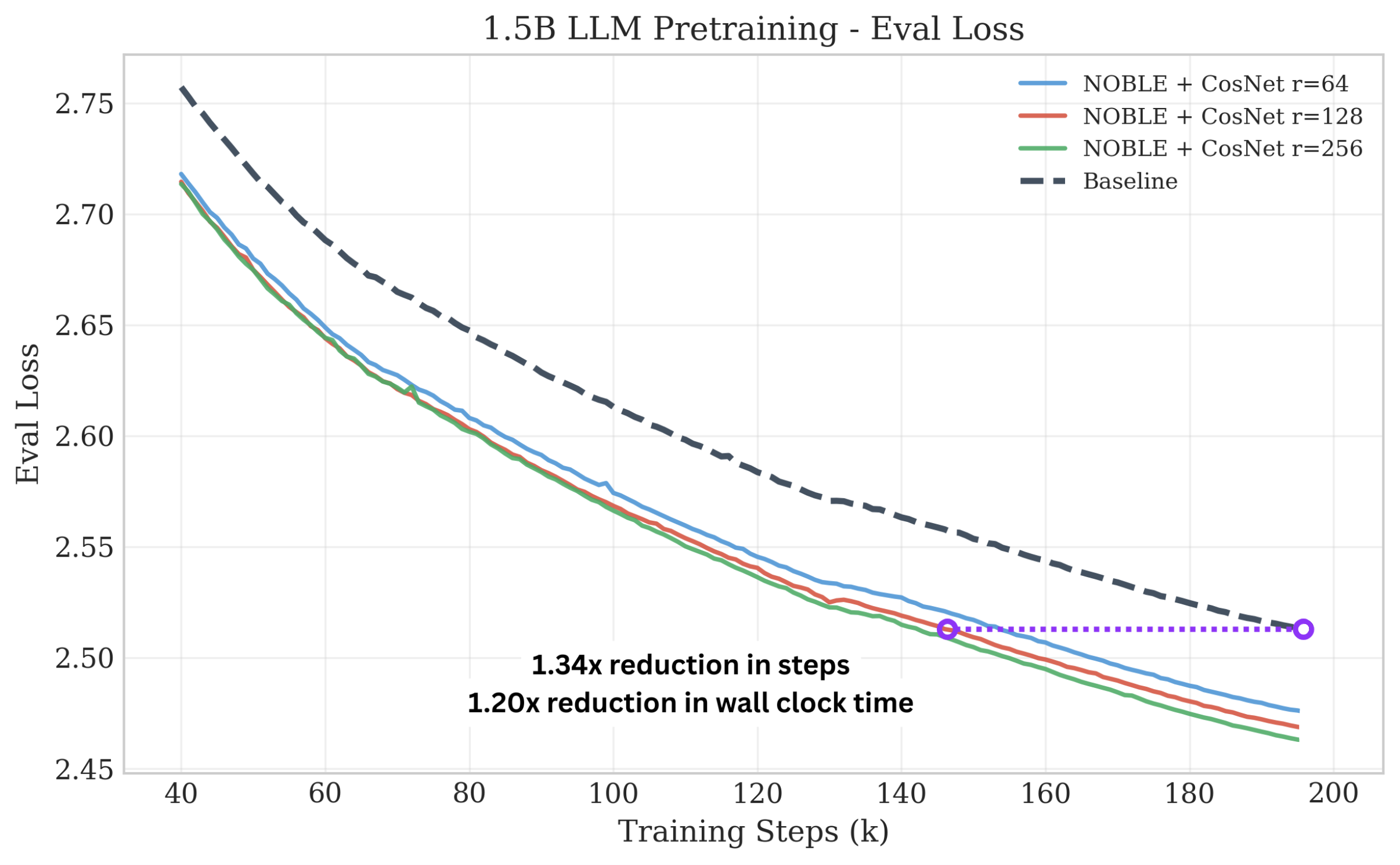}
\end{minipage}
\caption{\textbf{Eval loss curves for autoregressive language modeling.} Left: LLM Base 250M. Right: LLM Large 1.5B. NOBLE configurations (solid) consistently achieve lower loss than baseline (dashed) throughout training. Higher ranks provide greater improvement. First 20\% of training truncated for clarity.}
\label{fig:llm_curves}
\end{figure}

\textbf{Results.} Table~\ref{tab:lm_results} shows that NOBLE consistently achieves lower final loss than the baseline. We report \emph{step speedup} (ratio of baseline steps to steps needed to reach baseline's final loss) and \emph{wallclock speedup} (accounting for per-step overhead). Figure~\ref{fig:llm_curves} shows the training curves.

\begin{table}[t]
\centering
\caption{Autoregressive language modeling results on OpenWebText. NOBLE reaches baseline loss significantly faster despite per-step overhead, yielding net wallclock savings.}
\label{tab:lm_results}
\resizebox{\textwidth}{!}{%
\begin{tabular}{llccccccc}
\toprule
Model & Configuration & Rank & Params+ & Step Time+ & Eval Loss & $\Delta$ Loss & Step Speedup & Wallclock Speedup \\
\midrule
\multirow{4}{*}{\shortstack{Base 250M\\(12L, 1024d)}} 
& Baseline & -- & -- & -- & 2.850 & -- & 1.00$\times$ & 1.00$\times$ \\
& NOBLE+CosNet & 64 & +5.7\% & +7.6\% & 2.810 & --0.040 & 1.26$\times$ & 1.17$\times$ \\
& NOBLE+CosNet & 128 & +11.6\% & +11.5\% & 2.798 & --0.052 & 1.35$\times$ & 1.21$\times$ \\
& NOBLE+CosNet & 256 & +24.1\% & +20.8\% & 2.780 & --0.070 & 1.47$\times$ & 1.22$\times$ \\
\midrule
\multirow{4}{*}{\shortstack{Large 1.5B\\(24L, 2048d)}} 
& Baseline & -- & -- & -- & 2.513 & -- & 1.00$\times$ & 1.00$\times$ \\
& NOBLE+CosNet & 64 & +4.0\% & +6.8\% & 2.476 & --0.037 & 1.27$\times$ & 1.19$\times$ \\
& NOBLE+CosNet & 128 & +8.2\% & +11.3\% & 2.468 & --0.045 & 1.34$\times$ & 1.20$\times$ \\
& NOBLE+CosNet & 256 & +16.6\% & +14.7\% & 2.463 & --0.050 & 1.37$\times$ & 1.19$\times$ \\
\bottomrule
\end{tabular}%
}
\end{table}

\textbf{Computational overhead.} On the Base model, absolute step times are 309ms (baseline), 332ms (r64), 344ms (r128), and 373ms (r256), measured on 8$\times$H100 with batch 64. Despite 7--21\% longer step times, the 21--32\% reduction in steps to reach baseline loss yields 1.17--1.22$\times$ net wallclock speedup.

\subsubsection{BERT-Style Masked Language Modeling}

\textbf{Setup.} We train a BERT-style transformer with bidirectional attention on OpenWebText using masked language modeling (15\% masking probability, standard 80/10/10 mask/random/unchanged split). Architecture: depth 12, width 768, 12 attention heads. Training: batch 64, lr $3 \times 10^{-4}$, 100k steps, $\beta_1=0.9$, $\beta_2=0.98$, weight decay 0.01. We use RoPE positional encodings (unlike standard BERT) to match our LLM setup.

\textbf{Results.} Table~\ref{tab:bert_results} shows NOBLE also improves BERT pretraining, though with more modest speedups than those observed with our LLM experiments.

\begin{table}[t]
\centering
\caption{BERT-style MLM results on OpenWebText (100k steps).}
\label{tab:bert_results}
\begin{tabular}{lcccc}
\toprule
Configuration & Rank & Eval Loss & $\Delta$ Loss & Step Speedup \\
\midrule
Baseline & -- & 1.414 & -- & 1.00$\times$ \\
NOBLE+CosNet & 64 & 1.393 & --0.021 & 1.10$\times$ \\
NOBLE+CosNet & 128 & 1.371 & --0.044 & 1.17$\times$ \\
NOBLE+CosNet & 256 & 1.351 & --0.064 & 1.26$\times$ \\
\bottomrule
\end{tabular}
\end{table}

\subsection{Image Model Experiments}

\subsubsection{ViT-S ImageNet Classification}

\textbf{Setup.} We train ViT-Small (22M parameters) on ImageNet-1k for 90 epochs with batch size 1024, image size 224, RandAugment, learning rate $3 \times 10^{-4}$, $\beta_1=0.9$, $\beta_2=0.98$, weight decay 0.05, and cosine annealing with 5 warmup epochs. We compare configurations with and without Mixup/CutMix augmentation.

\begin{figure}[h]
\centering
\includegraphics[width=\textwidth]{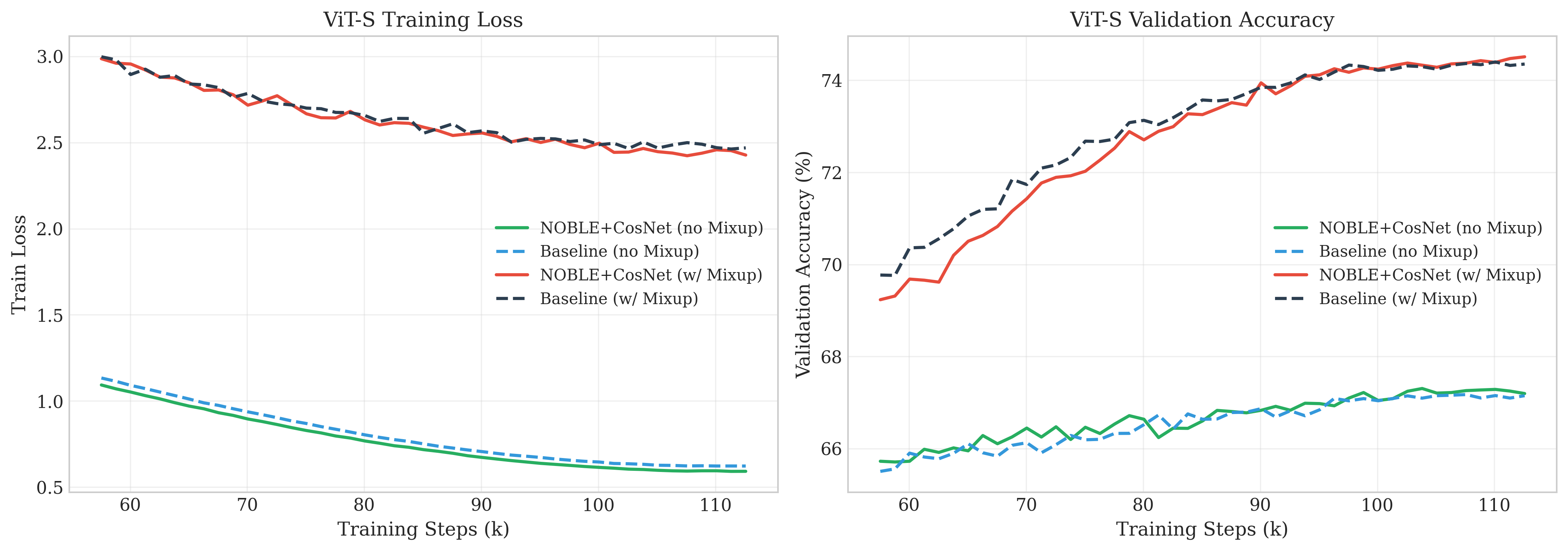}
\caption{ViT-S training loss and validation accuracy on ImageNet-1k (second half of training shown). \textbf{Left:} Training loss. Without Mixup/CutMix, NOBLE achieves 5\% lower training loss. Meanwhile when training with Mixup/CutMix, it is not clear that NOBLE provides any benefit. \textbf{Right:} Validation accuracy. Mixup/CutMix significantly boosts accuracy (67\% $\rightarrow$ 74\%), NOBLE does not provide much improvement in either condition.}
\label{fig:vit_results}
\end{figure}

\textbf{Results.} Table~\ref{tab:vit_results} and Figure~\ref{fig:vit_results} reveal an interesting interaction between NOBLE and augmentation. With standard Mixup/CutMix augmentation, NOBLE provides a modest improvement (1.6\% lower train loss). However, \textbf{without Mixup/CutMix, NOBLE's benefit is more pronounced (5.0\% lower train loss)}, suggesting that the aggressive label smoothing from these augmentations partially interferes with NOBLE's fitting capabilities.

\begin{table}[t]
\centering
\caption{ViT-S ImageNet-1k classification results. NOBLE benefits training when Mixup/CutMix augmentation is disabled.}
\label{tab:vit_results}
\begin{tabular}{lccc}
\toprule
Configuration & Mixup/CutMix & Train Loss & Top-1 Acc. \\
\midrule
Baseline & \checkmark & 2.462 & 74.40\% \\
NOBLE+CosNet (r=64) & \checkmark & 2.423 & 74.51\% \\
\midrule
Baseline & \texttimes & 0.622 & 67.17\% \\
NOBLE+CosNet (r=64) & \texttimes & 0.591 & 67.31\% \\
\bottomrule
\end{tabular}
\end{table}

\subsubsection{Autoregressive Image Token Modeling}
\label{sec:image_token}

\textbf{Setup.} We train an autoregressive transformer on discrete image tokens, analogous to LLM training on text tokens. Images are first encoded into discrete tokens using a pretrained VQGAN~\citep{esser2021taming}, then we train a causal transformer to predict tokens autoregressively (next-token prediction):
\begin{itemize}
    \item Architecture: depth 12, width 1024, 8 heads (identical to base LLM)
    \item Tokenizer: amused-512 VQGAN with 8192 codebook, 1024 tokens per 512$\times$512 image (32$\times$32 grid)
    \item Positional encoding: 2D RoPE (analogous to 1D RoPE for text)
    \item Attention: Causal (raster-scan token order)
    \item Training: batch 64, lr $3 \times 10^{-4}$, $\beta_1=0.9$, $\beta_2=0.98$, weight decay 0.01, 250k steps
    \item Regularization: None (no dropout, no augmentation)
    \item Class conditioning: Prepended class embedding
\end{itemize}

\textbf{Results.} Table~\ref{tab:image_token_results} shows that NOBLE improves autoregressive image token modeling, achieving lower eval loss than baseline. This setup is structurally identical to LLM pretraining (next-token prediction with cross-entropy loss), differing only in the token vocabulary (discrete image codes vs. text BPE tokens). Combined with our ViT results (which show improvement when Mixup/CutMix is disabled), this demonstrates that NOBLE benefits both language and image token prediction when aggressive augmentation is not used.

\begin{table}[t]
\centering
\caption{Autoregressive image token modeling on ImageNet. NOBLE improves next-token prediction on discrete image tokens, matching the pattern observed for language models.}
\label{tab:image_token_results}
\begin{tabular}{lccc}
\toprule
Configuration & Rank & Eval Loss & $\Delta$ Loss \\
\midrule
Baseline & -- & 6.723 & -- \\
NOBLE+CosNet & 64 & 6.699 & --0.024 \\
NOBLE+CosNet & 128 & 6.690 & --0.033 \\
NOBLE+CosNet & 256 & 6.682 & --0.041 \\
\bottomrule
\end{tabular}
\end{table}

\subsection{Summary}

The pattern that emerges is clear: \textbf{NOBLE improves training in low noise generative and classification tasks}. The only configuration where NOBLE shows no improvement is ViT with Mixup/CutMix enabled. When we disable these augmentations, ViT also benefits from NOBLE in train loss, though worsens eval loss as it contributes to overfitting.

\subsection{Ablation Studies}

We conduct ablations on the base language model (depth 12, width 1024, batch 32) trained for 150k steps.

\begin{figure}[h]
\centering
\includegraphics[width=0.85\textwidth]{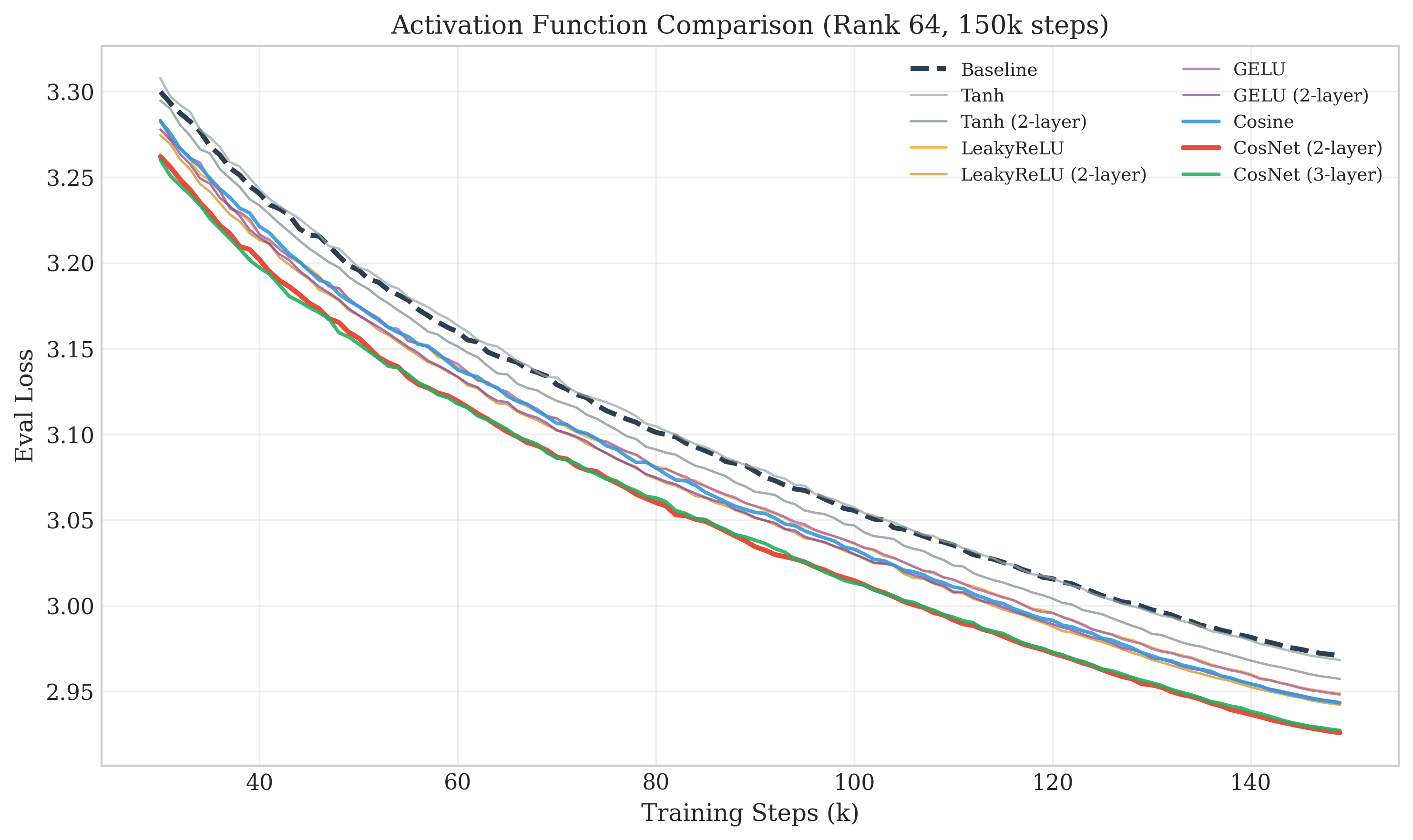}
\caption{\textbf{Activation function comparison.} Eval loss curves for different bottleneck activations (rank 64, 150k steps). CosNet (2-layer cosine, red) achieves the lowest loss, followed by single cosine and 3-layer CosNet. ReLU-like activations (GELU, LeakyReLU) provide moderate improvement, while tanh shows minimal benefit. First 20\% of training truncated for clarity.}
\label{fig:ablation_act}
\end{figure}

\paragraph{Activation Function and Depth Comparison.} Table~\ref{tab:ablation_act} and Figure~\ref{fig:ablation_act} compare different bottleneck activations, both as single activations and with a middle mixing projection $M$ (2-layer).

ReLU-like activations (LeakyReLU, GELU) provide moderate improvement. Tanh shows minimal benefit as a single activation but improves with the middle projection. Cosine consistently achieves the best results across both configurations. The 2-layer CosNet (with $M$) provides the best overall performance, while extending to 3 layers yields no additional improvement.

\begin{table}[h]
\centering
\caption{Ablation on bottleneck activation function (rank 64, base LLM, 150k steps).}
\label{tab:ablation_act}
\begin{tabular}{lcccc}
\toprule
Activation & Symmetric & Single Layer & 2-Layer (with $M$) & $\Delta$ from Baseline \\
\midrule
Baseline & -- & 2.971 & -- & -- \\
Tanh & \checkmark & 2.968 & 2.957 & --0.003 / --0.014 \\
LeakyReLU & \texttimes & 2.949 & 2.942 & --0.022 / --0.029 \\
GELU & \texttimes & 2.948 & 2.944 & --0.023 / --0.027 \\
Cosine & \checkmark & 2.943 & \textbf{2.926} & --0.028 / \textbf{--0.045} \\
\midrule
Cosine (3-layer) & \checkmark & -- & 2.927 & --0.044 \\
\bottomrule
\end{tabular}
\end{table}

\section{Discussion}

\paragraph{Why Cosine Activations?}

We evaluated several activation functions before settling on cosine (see Table~\ref{tab:ablation_act}). ReLU-like activations (ReLU, GELU) showed lower magnitudes of improvement. We hypothesize this relates to symmetry and the low-rank setting: ReLU-likes are asymmetric around zero, which may be problematic when projecting to \emph{smaller} dimensions. In standard MLPs, we project to \emph{higher} dimensions before applying ReLU, which acts as a selective gate, blocking certain activations and ``deleting'' information by collapsing it to zero. This filtering is often desirable for compression. In a low-rank bottleneck, this gating behavior may be counterproductive since we cannot afford to ``waste'' dimensions.

Symmetric activations like $\tanh$ address the asymmetry issue but suffer from saturation: much of the input range collapses to values near $\pm 1$, losing gradient signal. Cosine is symmetric around zero, bounded (avoiding exploding activations), and crucially \emph{non-saturating}. Its derivative $-\sin(x)$ oscillates rather than vanishing for large inputs.

Notably, periodic activations like sine and cosine are among the \emph{most nonlinear} options available, capable of aggressively fitting complex functions. This property underlies their success in SIREN~\citep{sitzmann2020implicit} and NeRF~\citep{mildenhall2020nerf}, where the goal is to overfit neural networks to specific signals or scenes. Empirical experiments on learning space-filling curves~\citep{smith2024spacefilling} further demonstrate that sinusoidal activations yield the strongest ability to densely cover output spaces. By pairing the main dense linear layer with a cosine-based bypass, we provide a branch that is maximally \emph{antithetical} to the linear pathway, potentially offering a unique nonlinear delta that cannot be achieved by any modification to the linear weights alone. In other words, the linear layer approximates the dominant, smoother low-frequency components, while the cosine-based bypass captures residual high-frequency variations with smaller amplitudes.

\paragraph{CosNet Architecture Evolution.}

Our initial experiments used a single cosine activation without the middle projection $M$. Even in this simple form, cosine outperformed other activations. We then explored whether adding depth to the bottleneck could further improve performance. Adding the mixing matrix $M \in \R^{r \times r}$ between two cosine activations (the CosNet architecture) yielded additional gains (see Table~\ref{tab:ablation_act}). Extending to three or more layers provided no further improvement but also did not hurt, suggesting two layers capture the useful complexity for this setting.

\paragraph{Function Smoothness and Regularization}

To understand NOBLE's interaction with augmentation, it helps to formalize what \textit{smoothness} means for a learned function. A function $f$ is $L$-Lipschitz if $\|f(x) - f(y)\| \leq L\|x - y\|$ for all $x, y$, bounding how rapidly outputs can change with respect to inputs. The related notion of $L$-smooth gradients requires $\|\nabla f(x) - \nabla f(y)\| \leq L\|x - y\|$~\citep{nesterov2004introductory}, bounding the rate of change of the gradient itself, i.e., constraining curvature. A function with a small gradient Lipschitz constant cannot exhibit sharp bends or rapid directional changes. In the frequency domain, this corresponds to energy concentrated in low-frequency Fourier components, while high-frequency components encode rapid oscillations, sharp transitions, and fine-grained local structure~\citep{rahaman2019spectral}.

Many standard regularization techniques implicitly or explicitly encourage such smoothness. Weight decay penalizes large weights, limiting each layer's Lipschitz constant and thus the composed function's capacity for rapid variation. Dropout averages over an implicit ensemble of sub-networks, producing predictions smoother than any individual member. Label smoothing~\citep{szegedy2016rethinking} softens targets, discouraging sharp decision boundaries. Spectral normalization~\citep{miyato2018spectral} directly bounds layer-wise Lipschitz constants. Gaussian noise injection is equivalent to Tikhonov regularization~\citep{bishop1995training, seghouane2004regularizing}, explicitly penalizing high-frequency components~\citep{camuto2020explicit}. Mixup~\citep{zhang2018mixup} and CutMix enforce that the network interpolates linearly between training points, imposing a direct smoothness constraint. These techniques are effective in part because neural networks exhibit a \emph{spectral bias}: they naturally learn low-frequency components first and fit high-frequency details only later in training, if at all~\citep{rahaman2019spectral}. Smoothness regularization amplifies this inductive bias to improve generalization.

\paragraph{NOBLE as a High-Frequency Residual Learner}

The target function generally contains both smooth, low-frequency components, the coarse global trend, and non-smooth, high-frequency components: fine-grained variations characterized by large higher-order derivatives, sharp bends, and rapid local changes. We hypothesize that NOBLE provides a \emph{division of labor}: the main linear pathway $xW$ captures the dominant smooth component, while the cosine bypass specializes in the high frequency residual that is not well-approximated by a smooth function. Cosine activations are naturally suited to this role because their learnable frequency parameters $\omega$ allow representation of arbitrary frequencies, and periodic activations are among the most aggressive function fitters available~\citep{sitzmann2020implicit}.

This framing explains why Mixup and CutMix specifically interfere with NOBLE. These augmentations do not merely encourage smoothness as a regularizer on the model. They modify the train-time target function itself by blending inputs and labels, making it inherently smoother~\citep{zhang2018mixup}. The high frequency residual that NOBLE's cosine branch is designed to capture is attenuated or removed. It is not that NOBLE fails, but that the augmentation eliminates the very structure NOBLE is designed to exploit. This is consistent with our observation that NOBLE improves training loss on every task except ViT ImageNet classification with Mixup/CutMix enabled, and even then, ViT image, does benefit when these augmentations are removed.

More broadly, practitioners should be aware that any regularization strategy that aggressively encourages smooth fits may reduce NOBLE's benefits by suppressing the high frequency signal the cosine branch learns. Tasks and training regimes that preserve fine-grained structure in the target function are where NOBLE provides the greatest advantage.

\paragraph{Relationship to Standard LoRA and PEFT}

Despite the structural similarity, NOBLE serves a fundamentally different purpose than LoRA and other PEFT methods. LoRA was designed to efficiently adapt \emph{frozen pretrained models} to downstream tasks by learning low-rank corrections. NOBLE, in contrast, is an \emph{architectural augmentation} for pretraining. The branch is a permanent part of the model trained from scratch, not an adapter attached to frozen weights. Where LoRA learns \emph{corrections}, NOBLE learns \emph{complementary features} that the main linear layer cannot represent.

\paragraph{Computational Tradeoffs}

At rank 64--128, NOBLE adds 4--12\% parameters and 7--12\% step time overhead, but provides 1.26--1.35$\times$ step speedup, yielding 1.17--1.21$\times$ net wallclock speedup. This represents a favorable tradeoff: practitioners can reach target performance significantly faster despite the per-step cost.

\paragraph{Future Directions}

We hope this work inspires further exploration into nonlinear branches for neural network architectures. The principle of combining a linear backbone with nonlinear residual branches for more granular fitting may extend beyond low-rank adaptations to other hybrid function parameterizations. Understanding when and why different nonlinearities complement linear pathways remains an open and promising research direction.

\section{Limitations}

\begin{itemize}
    \item \textbf{Inference overhead}: Unlike adapter methods that can be merged post-training, NOBLE adds permanent inference cost (6--12\% FLOPs). This may be undesirable for inference-constrained deployments.
    \item \textbf{Augmentation interaction}: We identify that Mixup/CutMix interferes with NOBLE, but do not fully characterize which augmentation strategies are compatible.
    \item \textbf{Limited vision coverage}: We test only ViT classification and autoregressive image token modeling. Other vision tasks (detection, segmentation) might behave differently.
    \item \textbf{Scale}: Our largest model is 1.5B parameters. The observed effects might change at larger scales.
    \item \textbf{Tokenizer dependence}: Our image token results depend on the specific VQGAN tokenizer (amused-512). Other visual tokenizers might yield different results.
\end{itemize}

\section{Conclusion}

We introduced NOBLE (Nonlinear lOw-rank Branch for Linear Enhancement), a family of nonlinear low-rank branches for transformer linear layers. Among several activation functions, we find CosNet, a two-layer cosine nonlinearity, performs best, achieving \textbf{up to 1.47$\times$ step speedup}. With 4--24\% additional parameters and 7--21\% per-step overhead, NOBLE reduces steps to reach baseline loss by 21--32\%, yielding \textbf{net wallclock speedups of 1.17--1.22$\times$}. This makes it a practical method to accelerate pretraining with minimal added complexity.

Our experiments demonstrate that NOBLE benefits a broad range of tasks: LLM autoregressive modeling, BERT masked language modeling, autoregressive image token modeling, and ViT image classification. The one caveat is an interaction with Mixup/CutMix augmentation: when these aggressive augmentations are used, NOBLE's benefits are reduced. When disabled, ViT also benefits from NOBLE, suggesting the conflict is with the augmentation strategy rather than the task or modality.

\bibliographystyle{plainnat}
\bibliography{references}

\appendix

\section{Implementation Details}

\paragraph{CosNet Architecture.} The full CosNet nonlinearity applies two cosine activations with a linear mixing layer:
\begin{align}
    h_1 &= \cos(\omega_1 \odot x + \phi_1) \\
    h_2 &= M h_1 \\
    \text{CosNet}(x) &= \cos(\omega_2 \odot h_2 + \phi_2)
\end{align}
where $M \in \R^{r \times r}$ is initialized with Xavier initialization and uses an elevated learning rate (scaled by the net lr mult power, default 0.45).

\paragraph{Frequency Initialization.} Frequencies are sampled uniformly from $[\omega_\text{min}, \omega_\text{max}] = [0.8, 1.2]$. Meanwhile phases are sampled from a normal distribution with mean 0 and standard deviation 0.1. Initializations with larger ranges and variances underperformed.

\paragraph{Hyperparameter Defaults.}
\begin{itemize}
    \item LoRA rank: 64--256
    \item Learning rate multiplier power $\gamma$: 0.3 ($W_\text{up}$ uses $2\gamma = 0.6$)
    \item $W_\text{up}$ initialization scale $\alpha$: 0.01
    \item Main linear $W$ initialization scale: $0.5 / \sqrt{d_\text{in}}$ (half typical Kaiming)
    \item Frequency learning rate multiplier: 3.0
    \item Phase learning rate multiplier: 5.0
    \item Phase initialization std: 0.1
    \item CosNet internal fc lr mult power: 0.45
\end{itemize}

\section{Experimental Setup Details}

\paragraph{LLM Training.} OpenWebText dataset, BPE tokenizer (50257 vocab), sequence length 1024, linear warmup for 2000 steps, linear decay.

\paragraph{BERT Training.} Same dataset, BERT tokenizer (30522 vocab), sequence length 1024, linear warmup for 1000 steps, linear decay.

\paragraph{ViT Training.} ImageNet-1k, 224$\times$224 images, ViT-S/16 architecture (patch size 16), RandAugment (n=2, m=9), Mixup ($\alpha$=0.8), CutMix ($\alpha$=1.0), linear warmup for 5 epochs followed by linear decay.

\paragraph{Image Token Modeling.} ImageNet-1k images encoded to 1024 discrete tokens (32$\times$32 grid) using pretrained amused-512 VQGAN (8192 codebook). Autoregressive transformer trained with next-token prediction (cross-entropy loss), class-conditional with prepended class embedding, 2D RoPE positional encoding, raster-scan token order.

\end{document}